\documentclass{article}

    \PassOptionsToPackage{numbers, compress}{natbib}

\usepackage[dblblindworkshop, final]{neurips_2026}
\workshoptitle{Reinforcement Learning for Experimental Sciences: Bridging the Simulation-to-Reality Gap}

\usepackage[utf8]{inputenc} 
\usepackage[T1]{fontenc}    
\usepackage{hyperref}       
\usepackage{url}            
\usepackage{booktabs}       
\usepackage{amsfonts}       
\usepackage{nicefrac}       
\usepackage{microtype}      
\usepackage{xcolor}         

\usepackage{amsmath,amssymb,amsthm,graphicx,multirow}
\usepackage{tikz}  
\usetikzlibrary{arrows.meta,positioning,calc,decorations.pathreplacing}

\newtheorem{proposition}{Proposition}
\newtheorem{corollary}{Corollary}
\newcommand{\framework}{\textsc{MD-BED}}
\newcommand{\mddad}{\textsc{MD-DAD}}
\newcommand{\mddadps}{\textsc{MD-DAD-PS}}
\newcommand{\indad}{\textsc{InDAD}}
\newcommand{\indadps}{\textsc{InDAD-PS}}
\newcommand{\mddadpsid}{\textsc{MD-DAD-PS-ID}}
\newcommand{\indadpsid}{\textsc{InDAD-PS-ID}}

\usepackage{booktabs}
\usepackage{tabularx}
\usepackage{makecell}
\usepackage{colortbl}

\newtheorem{definition}{Definition}

\title{MD-BED: Coordinating Multiple Designers in Bayesian Experimental Design}

\author{%
  Marcel Hedman \\
  Department of Statistics\\
  University of Oxford\\
  \texttt{marcel.hedman@jesus.ox.ac.uk} \\
}

\hypersetup{
  pdftitle={MD-BED: Coordinating Multiple Designers in Bayesian Experimental Design},
  pdfauthor={Marcel Hedman},
}

\begin{document}

\maketitle

\begin{abstract}
\looseness=-1
To date, Bayesian experimental design (BED) has largely studied how a single adaptive policy should choose experiments. Yet many real-world data-acquisition systems distribute experimental decisions across multiple designers, such as laboratories, field teams, or decentralised robots. We introduce multi-designer Bayesian experimental design (MD-BED) as a general paradigm for BED in which experimental decisions are distributed across multiple designers, and within this paradigm focus on a fully cooperative setting with decentralised execution. In this regime, designers share an inferential objective but choose experiments using only their local histories, creating a coordination problem: individually informative policies may collectively acquire redundant information. We formalise this failure through an information-theoretic coordination gap and show that, under conditionally independent acquisition, it is exactly characterised by the total correlation among designer histories. To address this problem, we adapt centralised training with decentralised execution (CTDE) to BED, jointly optimising locally executable policies against the information contained in their pooled experimental histories. We instantiate this framework with DAD-based algorithms trained using a contrastive information objective. Empirically, joint training reduces coordination failure, induces greater designer specialisation under heterogeneous experimental costs, and outperforms independently trained policies without requiring communication at execution. These results establish coordination as an important challenge in Bayesian experimental design when experimental decisions are distributed across multiple designers.
\end{abstract}

\section{Introduction}
\label{sec:intro}

Intelligent, adaptive data acquisition is central to scientific discovery and is increasingly important in settings where decision-makers must determine what information to gather under limited resources \citep{mackay1992information, atkinson2007optimum}. Bayesian experimental design (BED) provides a principled framework for this problem, selecting experiments according to the information they are expected to provide about unknown quantities of interest \citep{lindley1956, chaloner1995, ryan2016review, rainforth2024modern}. Within the field, the canonical objective is the expected information gain (EIG), which measures the expected reduction in uncertainty from prior to posterior \citep{lindley1956, shannon1948mathematical}. Let $\theta$ denote an unknown quantity of interest, $\xi$ an experimental design, and $y$ an observation generated according to $p(y\mid\theta,\xi)$. The EIG of $\xi$ is
\begin{equation}
\operatorname{EIG}(\xi)
= I(\theta;y\mid\xi)
= \mathbb{E}_{p(\theta)p(y\mid\theta,\xi)}
\left[
\log\frac{p(y\mid\theta,\xi)}{p(y\mid\xi)}
\right],
\label{eq:myopic_eig}
\end{equation}
where $p(y\mid\xi)=\int p(y\mid\theta,\xi)p(\theta)\mathrm{d}\theta$. BED therefore evaluates an experiment by the information it is expected to provide before its outcome is known.

Beyond this single-step setting, BED is often performed sequentially, allowing later experiments to adapt to observations collected earlier. Traditionally, this entails repeatedly updating the posterior and re-optimising the next design myopically, which can make adaptive experimentation expensive at deployment. Recent policy-based methods instead amortise this process by learning mappings from experimental histories $H_t$ directly to future designs, enabling rapid deployment-time decisions and non-myopic optimisation over complete experimental trajectories \citep{foster2021dad, ivanova2021implicit, hedman2025stepdad, Huang2025ALINEJA, rossa2026actionbedtaskdrivenbayesianexperimental}. BED is consequently becoming a progressively broader framework for adaptive information acquisition, with recent applications ranging from likelihood-free scientific experiments to information gathering with large language models and scientific model discovery \citep{choudhury2026bedllm,Murphy2026ModelDA}.

One structural assumption, however, remains largely implicit across these developments in the literature: experimental decisions are typically treated as belonging to a \emph{single designer}. In practice many real data-acquisition systems instead distribute experimental capacity across multiple designers. Separate laboratories may conduct experiments toward a common scientific objective \citep{bai_mosbach_taylor_karan_lee_rihm_akroyd_lapkin_kraft_2024}, field teams may independently choose where to measure, distributed trial sites may adapt local interventions, and teams of robots may probe different parts of an unknown system. Related structures do appear across batch experimental design, sensor placement, decentralised active information gathering, Dec-POMDPs, and multi-robot active search \citep{ma_info_gathering,pmlr-v161-ghods21a,Gupta-2026-88364}. These literatures establish that multiple decision-makers can acquire information jointly, but they approach the problem using machinery from distinct fields, are domain specific, and/or do not develop multi-designer experimental design as a general organising perspective within BED itself. 

\begin{figure}[t]
\centering

\definecolor{band}{HTML}{E6E6E6}
\definecolor{blueF}{HTML}{E4EEFB}   \definecolor{blueM}{HTML}{C3D9F5}   \definecolor{blueD}{HTML}{2F6BB0}
\definecolor{orangeF}{HTML}{FDEBDD} \definecolor{orangeM}{HTML}{F8CBA6} \definecolor{orangeD}{HTML}{D9712C}
\definecolor{greenF}{HTML}{E6F3E3}  \definecolor{greenM}{HTML}{C6E5BF}  \definecolor{greenD}{HTML}{3D8A48}

\tikzset{
  >={Stealth[length=2.3mm,width=1.7mm]},
  arr/.style={->,line width=1pt},
  step/.style={draw,rounded corners=4pt,line width=0.9pt,minimum width=#1,
               minimum height=0.5cm,inner sep=2pt,font=\bfseries},
  step/.default=2.15cm,
  pol/.style={circle,draw,line width=0.9pt,minimum size=0.68cm,inner sep=0pt},
  hist/.style={draw,rounded corners=4pt,line width=0.9pt,minimum width=0.85cm,
               minimum height=0.55cm,inner sep=2pt},
  lab/.style={font=\footnotesize},
}

\newcommand{\panel}[5]{%
  \begin{scope}
    \clip[rounded corners=6pt] (#1,#2) rectangle (#3,#4);
    \fill[white] (#1,#2) rectangle (#3,#4);
    \fill[band]  (#1,{#4-0.55}) rectangle (#3,#4);
  \end{scope}
  \draw[rounded corners=6pt,line width=0.8pt,black!75] (#1,#2) rectangle (#3,#4);
  \node[anchor=west,font=\bfseries] at ({#1+0.12},{#4-0.275}) {#5};
}

\def\stepw{2.5cm}
\newcommand{\loopglyph}[7]{%
  \node[step=\stepw,fill=#4M,draw=#4D,align=center,inner sep=3pt] (D) at (#1,#2)
       {Design\\[1pt]\normalfont\small$#5$};
  \node[step=\stepw,fill=#4M,draw=#4D,align=center,inner sep=3pt] (O) at (#1,#3)
       {Outcome\;\, \normalfont\small$#6$};
  \draw[arr] (D.east) -- ({#1+#7},#2) -- ({#1+#7},#3) -- (O.east);
  \draw[arr] (O.west) -- ({#1-#7},#3) -- ({#1-#7},#2) -- (D.west);
  \node[font=\small\itshape] at ($(D.south)!0.5!(O.north)$) {$T$ rounds};
}

\newcommand{\designer}[3]{%
  \node[pol,fill=#3M,draw=#3D] (p#2) at (9.75,#1) {$\pi_{#2}$};
  \node[hist,fill=#3F,draw=#3D] (h#2) at (11.6,#1) {$H_{#2}$};
  \draw[arr] (p#2.east) -- (h#2.west);
}

\resizebox{\linewidth}{!}{%
\begin{tikzpicture}

\panel{0}{0}{4.1}{5.2}{(a) Single-designer BED}
\def\stepw{2.7cm}\loopglyph{2.05}{3.2}{1.4}{blue}{\xi_t\sim\pi(\cdot\,|\,H_{t-1})}{y_t}{1.72}

\begin{scope}[xshift=0.15cm]
\panel{4.15}{0}{9.1}{5.2}{(b) Multi-designer BED}
\foreach \c/\d in {green/0.3,orange/0.15}
  \fill[\c F,draw=\c D,rounded corners=5pt,line width=0.8pt]
       ({4.75+\d},{0.35+\d}) rectangle ({8.62+\d},{4.0+\d});
\fill[blueF,draw=blueD,rounded corners=5pt,line width=0.8pt] (4.75,0.35) rectangle (8.62,4.0);
\node[anchor=west,font=\bfseries] at (4.85,3.68) {Designer $\pi_i$};
\def\stepw{2.95cm}\loopglyph{6.68}{2.78}{1.1}{blue}{\xi_{i,t}\sim\pi_i(\cdot\,|\,\mathcal{F}_{i,t-1})}{y_{i,t}}{1.78}
\node[font=\small] at (4.46,2.15) {$N{\times}$};
\end{scope}

\begin{scope}[xshift=0.95cm]
\panel{8.5}{0}{17.3}{5.2}{(c) Cooperative, decentralised \framework}
\node[font=\bfseries\small,text=black!65] at (10.68,4.22) {Acquired locally};
\node[font=\bfseries\small,text=black!65] at (15.45,4.22) {Valued jointly};
\draw[dashed,black!30,line width=0.6pt] (12.95,0.3) -- (12.95,4.4);

\designer{3.45}{1}{blue}
\designer{2.5}{2}{orange}
\node at (9.75,1.62) {$\vdots$};
\node at (11.6,1.62) {$\vdots$};
\designer{0.65}{N}{green}

\draw[decorate,decoration={brace,amplitude=6pt},line width=0.9pt]
     (12.22,3.75) -- (12.22,0.35);
\node[draw=greenD,fill=greenM,rounded corners=4pt,line width=1.1pt,align=center,
      inner xsep=3.5pt,inner ysep=6pt,minimum width=2.55cm] (peig) at (15.45,2.05)
      {\bfseries PEIG$(\boldsymbol{\pi})$\\[3pt]$\max\limits_{\boldsymbol{\pi}}\, I(\theta;\mathbf{H}_{1:N}\mid\boldsymbol{\pi})$};
\draw[arr] (12.5,2.05) -- node[above,lab,xshift=0.25cm]{$\mathbf{H}_{1:N}$} (peig.west);
\end{scope}

\end{tikzpicture}%
}
\caption{From single-designer to multi-designer Bayesian experimental design (\framework).
\textbf{(a)} Sequential BED involves a single designer, whose policy $\pi$
selects each design from the history $H_{t-1}$; for example, $\pi$ may be an amortised network
or a posterior update followed by design optimisation.
\textbf{(b)} \framework\ generalises this setting by distributing experimental decisions across $N$
designers, where designer $\pi_i$ acts on its available information $\mathcal{F}_{i,t-1}$, which can
include, among other things, its own history $H_{i,t-1}$ or the joint history $\mathbf{H}_{1:N,t-1}$.
Designers may also
differ in their objectives, capabilities and outcome models
(Table~\ref{tab:mdbed-settings} in Appendix~\ref{app:mdbed-settings} gives a taxonomy of these settings).
\textbf{(c)} In the cooperative, decentralised setting considered here, each designer $\pi_i$
acquires data using only its own local history, but all designers share a single objective defined
on the pooled history $\mathbf{H}_{1:N}$: the pooled expected information gain (PEIG). Other
objectives, such as task-driven ones \citep{rossa2026actionbedtaskdrivenbayesianexperimental},
could also be used.}
\label{fig:overview}
\end{figure}
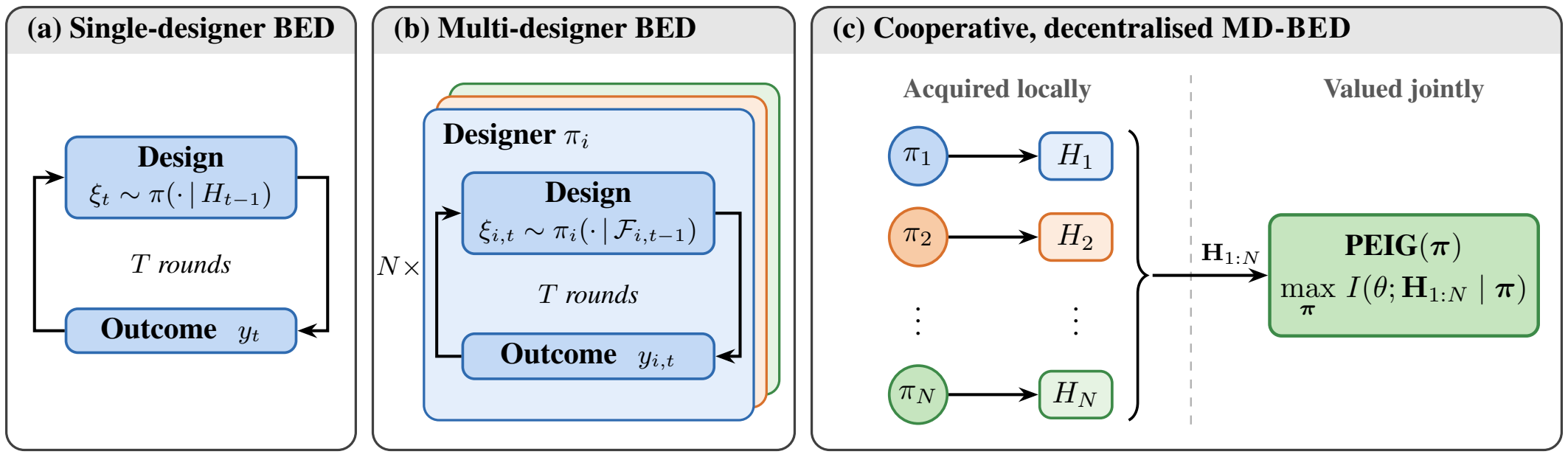

We therefore introduce \emph{multi-designer Bayesian experimental design} (\framework; Figure~\ref{fig:overview}) as a paradigm for BED in which experimental decisions are distributed across multiple designers. This viewpoint encompasses a range of possible interaction structures: designers may be cooperative, competitive, or mixed-motive; execution may be centralised or decentralised; interaction may be synchronous, asynchronous, or organised into rounds; and designers may differ in their experimental capabilities, observation processes, constraints, and costs. In this work, we focus on one practically important regime: \emph{fully cooperative designers with decentralised execution}, where all designers seek information about a shared latent quantity, while at test-time each chooses experiments using only its own local experimental history.

This distinction between \emph{acquiring locally} and \emph{valuing jointly} is central to our formulation. Each designer may have its own design space and acquisition model, yet the experiments ultimately contribute to a shared inferential task. We therefore evaluate a collection of designer policies through the information contained in their pooled experimental histories, rather than by summing the information that each designer acquires in isolation. This immediately exposes a difficulty that does not arise for a single experimental decision-maker: \emph{coordination}. Two designers can each select experiments that are highly informative individually while collectively gathering largely the same information. More generally, independently optimising each designer's acquisition policy can reward information already expected to be obtained elsewhere in the team. Scaling BED from one designer to many therefore introduces a coordination problem in addition to familiar problems like adaptivity.

\looseness=-1
We make this failure precise through an information-theoretic \emph{coordination gap}, defined by the discrepancy between summed designer-wise information and the information in the pooled dataset. Under conditionally independent acquisition, we show that this gap is exactly characterised by the predictive total correlation among designer histories, providing a direct information-theoretic measure of redundant acquisition. This characterisation motivates leveraging machinery from cooperative multi-agent reinforcement learning (MARL) \citep{Buoniu2010MultiagentRL}. Specifically, we use \emph{centralised training with decentralised execution} (CTDE), in which information unavailable during deployment may nevertheless be used to coordinate policies during learning \citep{lowe2017multiagent,rashid2018qmix,hedman2026coda,mappo,foerster2018coma}. We adapt this principle from MARL to BED. During training, we simulate the complete multi-designer experiment and optimise all designer policies jointly against the information in the pooled history; during execution, each designer remains restricted to its own local observations. Centralisation therefore enters only through the training objective, allowing a policy to learn that an experiment is valuable not merely because it is informative in isolation, but because it complements what the rest of the team is expected to acquire.

We operationalise this framework by extending Deep Adaptive Design (DAD) \citep{foster2021dad} to coordinated multi-designer acquisition. Our resulting DAD-based algorithms retain factorised, locally executable policies while training them against a shared information objective defined on the pooled experiment. We additionally allow designers to have heterogeneous experimental costs, in which case coordination determines not only \emph{what} information should be acquired but \emph{which designer} should acquire it. Empirically, joint training substantially increases the information in the pooled dataset relative to independently trained policies, reduces the coordination gap, and induces complementary acquisition roles. These benefits persist as the number of designers increases, can emerge under shared policy parameters when designers can distinguish their roles, and extend naturally to heterogeneous cost structures.

Taken together, our contributions develop multi-designer BED (\framework) as a distinct perspective within Bayesian experimental design. Rather than claiming that cooperative information gathering by multiple agents is itself new, we place such problems explicitly within BED; introduce a sub-setting in which locally acquired experiments are evaluated through a shared inferential objective while acting in a decentralised manner; provide an exact information-theoretic characterisation of coordination failure; adapt CTDE to BED as a mechanism for addressing this failure; and develop DAD-based algorithms that realise this approach in practice. More broadly, \framework ~provides a foundation for studying experimental design when information acquisition is distributed across multiple decision-makers rather than controlled by one.

\section{Background}
\label{sec:background}

\subsection{Single-Designer Bayesian Experimental Design}
\label{sec:background_bed}

Single-designer BED provides a framework for efficient and rational data acquisition in budget-constrained settings \citep{lindley1956}. As described in Section~\ref{sec:intro}, standard BED selects a design according to the information it is expected to provide about an unknown quantity of interest (Eq.~\eqref{eq:myopic_eig}).

In sequential BED, newly acquired observations are incorporated into the posterior before selecting the next experiment. Let $H_t=((\xi_1,y_1),\ldots,(\xi_t,y_t))$ denote the history after $t$ rounds. Conventional adaptive BED updates $p(\theta\mid H_{t-1})$ after observing $y_t$ and then optimises the next design under this updated belief \citep{ryan2016review}. While this produces an adaptive strategy, the resulting decisions are typically myopic: each experiment is selected for its immediate expected utility without explicitly accounting for how its possible outcomes affect later design choices.

Non-myopic approaches instead optimise the complete adaptive experiment \citep{foster2021dad}. A design policy $\pi$ maps the current history to the next design, $\xi_t\sim\pi(\cdot \mid H_{t-1})$. Writing $H:=H_T$ for the complete random history generated by executing $\pi$, the non-myopic trajectory-level information objective is
\begin{equation}
\mathcal{I}_T(\pi)
=
\mathbb{E}_{p(\theta,H\mid\pi)}
\left[
\log
\frac{p(H\mid\theta,\pi)}
     {p(H\mid\pi)}
\right]
=
I(\theta;H\mid\pi),
\label{eq:trajectory_EIG}
\end{equation}
where, for a realised adaptive history,
$
p(H|\theta, \pi)
:=
\prod_{t=1}^{T}
\pi(\xi_{t}\mid H_{t-1})p(y_t\mid\theta,\xi_t,H_{t-1}).
$
Since the policy terms do not depend on $\theta$, they cancel in the likelihood ratios appearing in the information objective and the contrastive bounds of Section~\ref{sec:background_estimators}. These quantities can therefore be evaluated using only the observation likelihoods, regardless of whether the policy is deterministic or stochastic. For deterministic policies, as used in our experiments, the conditional distribution over designs $\pi(\xi_t \mid H_{t-1})$ is a point mass rather than a continuous density. \citet{foster2021dad} therefore omit the policy terms and write the trajectory likelihood as $\prod_{t=1}^{T} p(y_t \mid \theta, \xi_t, H_{t-1})$, which is formally a density over the observation sequence, from which the complete history can be reconstructed under the fixed deterministic policy. We retain the full form above for generality; since the policy terms cancel, both conventions give identical objectives and bounds.
Policy-based BED amortises this optimisation by learning the mapping from histories to designs before deployment. Deep Adaptive Design (DAD) \citep{foster2021dad} parameterises $\pi$ with a neural network and optimises it directly against a trajectory-level information objective, enabling adaptive design decisions by a forward pass at deployment. Related approaches extend this idea to implicit likelihoods and alternative amortisation schemes \citep{ivanova2021implicit, hedman2025stepdad, Huang2025ALINEJA}.
Throughout, we allow history-dependent observation models of the form $y_t\sim p(y_t\mid\theta,\xi_t,H_{t-1})$, permitting earlier interventions to alter the system on which later observations are made. 

\subsection{Contrastive EIG Estimation for Single-Designer BED}
\label{sec:background_estimators}

\looseness=-1
The EIG objectives (Eq.~\eqref{eq:myopic_eig} and~\eqref{eq:trajectory_EIG}) are generally intractable because of difficulty evaluating the marginal history density $p(H\mid\pi)$. A rich literature has therefore arisen for estimating the EIG itself \citep{foster2020unified, foster2019variational, pmlr-v80-rainforth18a, cnotrastive_iollo}. 
In particular, DAD typically uses sequential prior contrastive estimation (sPCE) to obtain a tractable lower bound for training \citep{foster2021dad}. 
 Let $\theta_0\sim p(\theta)$ generate a trajectory $H$, and draw $L$ additional contrastive latents $\theta_{1:L}\stackrel{\mathrm{iid}}{\sim}p(\theta)$. Then
\begin{equation}
\mathcal{L}_{\mathrm{PCE}}(\pi;L)
=
\mathbb{E}
\left[
\log
\frac{p(H|\theta_0, \pi)}
{\frac{1}{L+1}\sum_{\ell=0}^{L}p(H|\theta_\ell, \pi)}
\right]
\leq
I(\theta;H\mid\pi).
\end{equation}
When the simulator and policy are reparameterisable, this objective can be differentiated through the full simulated experiment and used to train $\pi$ end-to-end. For evaluation, sequential nested Monte Carlo (sNMC) gives a corresponding upper bound,
\begin{equation}
\mathcal{U}_{\mathrm{NMC}}(\pi;L)
=
\mathbb{E}
\left[
\log
\frac{p(H|\theta_0, \pi)}
{\frac{1}{L}\sum_{\ell=1}^{L}p(H|\theta_\ell, \pi)}
\right]
\geq
I(\theta;H\mid\pi).
\end{equation}
Together, sPCE and sNMC provide lower and upper estimates of the information acquired by a policy. Other variational and likelihood-free estimators can be used in settings where these assumptions do not hold \citep{ivanova2021implicit}.

\section{Multi-Designer Bayesian Experimental Design}
\label{sec:method}

To enable a clear exposition of BED with multiple designers, it is useful to introduce an overarching paradigm for this class of problems. We term this paradigm \emph{multi-designer Bayesian experimental design} (\framework), and use it to refer broadly to BED settings in which experimental decisions are distributed across multiple designers. Depending on the application, designers may differ in their objectives, information access, capabilities, observation processes, constraints, and costs, and may operate under centralised, decentralised, cooperative, or competitive interaction structures. Table~\ref{tab:mdbed-settings} in Appendix~\ref{app:mdbed-settings} summarises several common settings encompassed by this perspective.

\subsection{Cooperative Decentralised Formulation}
\label{sec:mdbed-formulation}

In this work, we focus on a fully cooperative, decentralised \framework~setting with a shared inferential target. Within this setting, we adopt a joint inferential objective: experiments are acquired locally but valued through the pooled history. 

Let $\theta\sim p(\theta)$ denote the latent quantity of interest and $\mathcal{I}:=\{1,\ldots,N\}$ a set of $N$ designers acquiring data over $T$ rounds. At round $t$, designer $i\in\mathcal{I}$ selects a design $\xi_{i,t}\in\mathcal{X}_i$ and receives a local observation $y_{i,t}\sim p_i(y\mid\theta,\xi_{i,t},H_{i,t-1})$, where $H_{i,t}:=((\xi_{i,s},y_{i,s}))_{s=1}^{t}$ is its local experimental history, with $H_{i,0}=\varnothing$ and $H_i:=H_{i,T}$. Dependence on $H_{i,t-1}$ allows earlier experiments by a designer to alter the system on which its later observations are made. We assume that a designer's outcomes do not additionally depend on the designs or histories of other designers.

We write $\mathbf{H}_{1:N,t}:=(H_{i,t})_{i\in\mathcal{I}}$ for the joint history, and $\boldsymbol{\xi}_t:=(\xi_{i,t})_{i\in\mathcal{I}}$ for the joint design. Each designer follows a local policy $\pi_i$, with $\boldsymbol{\pi}:=(\pi_i)_{i\in\mathcal{I}}$ denoting the corresponding joint policy. In general, designer $i$ selects $\xi_{i,t}\sim\pi_i(\cdot\mid\mathcal{F}_{i,t-1})$, where $\mathcal{F}_{i,t-1}$ is the information available to it before round $t$, which can include, among other things, its own history $H_{i,t-1}$ or the joint history $\mathbf{H}_{1:N,t-1}$ (Table~\ref{tab:mdbed-settings}). Under decentralised execution, $\mathcal{F}_{i,t-1}=H_{i,t-1}$, so that
\begin{equation}
\boldsymbol{\pi}(\boldsymbol{\xi}_t\mid\mathbf{H}_{1:N,t-1})
=
\prod_{i\in\mathcal{I}}
\pi_i(\xi_{i,t}\mid H_{i,t-1}).
\label{eq:joint-policy}
\end{equation}
 Eq.~\eqref{eq:joint-policy} is the key execution constraint: the team may be trained jointly, but each designer acts only from information available locally.

\looseness=-1
Under this formulation, data are \emph{acquired locally but valued jointly}. Each designer generates its own experimental history, while the resulting data contribute to a shared inferential task. We therefore evaluate the team through the information contained in the joint experimental history, rather than by summing the information acquired by each designer in isolation. This choice is specific to the setting considered here: cooperative \framework\ is defined only by the presence of a shared team objective, and does not in general require either a shared inferential target or an objective defined on the pooled history. In particular, summing designer-wise information would still fall within the definition of cooperative \framework, and may even be appropriate when designers target distinct latent quantities. When the designers instead acquire data about a common latent target, as here, evaluating the pooled history provides a natural measure of the information acquired by the team. This leads to Definition \ref{def:peig}.

\begin{definition}[Pooled Expected Information Gain]
\label{def:peig}
For a joint policy $\boldsymbol{\pi}$ inducing terminal joint history
$\mathbf{H}_{1:N}$ across multiple designers, we define the \emph{pooled expected information gain}
(PEIG) as
\begin{equation}
\operatorname{PEIG}(\boldsymbol{\pi})
:=
I(\theta;\mathbf{H}_{1:N}\mid\boldsymbol{\pi})
=
\mathbb{E}
\left[
\log
\frac{p(\mathbf{H}_{1:N}\mid\theta,\boldsymbol{\pi})}
     {p(\mathbf{H}_{1:N}\mid\boldsymbol{\pi})}
\right].
\label{eq:peig}
\end{equation}
\end{definition}

PEIG evaluates all data pooled across designers once; information acquired
by several designers contributes only through the additional information it provides
to the joint history.
Combining the local observation models with Eq.~\eqref{eq:joint-policy} gives
\begin{equation}
p(\mathbf{H}_{1:N}\mid\theta,\boldsymbol{\pi})
=
\prod_{i\in\mathcal{I}}\prod_{t=1}^{T}
\pi_i(\xi_{i,t}\mid H_{i,t-1})\,
p_i(y_{i,t}\mid\theta,\xi_{i,t},H_{i,t-1})
=
\prod_{i\in\mathcal{I}}p(H_i\mid\theta,\pi_i).
\label{eq:md-history-factorisation}
\end{equation}
Thus the complete local histories are conditionally independent across designers given $\theta$ under the acquisition structure studied here, even when a designer's outcomes depend on its own past.

\subsection{Coordination Failure in Multi-Designer BED}
\label{subsec:coordination}

The cooperative, decentralised setting above naturally raises a question of
\emph{coordination}: how can locally acting designers collectively acquire the most useful
information about the shared inferential target $\theta$? Intuitively, designers optimised
independently should coordinate less effectively than designers whose policies can exploit joint
information during training. In particular, while independently optimising each designer to maximise the EIG of its
own history (i.e. $\sum_{i\in\mathcal{I}} I(\theta;H_i\mid\pi_i)$) preserves decentralised execution and reuses standard BED machinery, it likely does not account for what the others are expected to learn. Information that several
designers acquire can therefore be rewarded multiple times, despite contributing only once to the
pooled inferential objective in Eq.~\eqref{eq:peig}.

To characterise this mismatch formally, we define a \emph{coordination gap} in Proposition \ref{prop:coordination-gap}. First, let the total
correlation among designer histories be defined as
\begin{equation}
\operatorname{TC}(\mathbf{H}_{1:N})
:=
\operatorname{KL}
\left(
p(\mathbf{H}_{1:N})
\,\middle\|\,
\prod_{i\in\mathcal{I}}p(H_i)
\right)
=
\sum_{i\in\mathcal{I}}\mathcal{H}(H_i)
-
\mathcal{H}(\mathbf{H}_{1:N}),
\label{eq:total-correlation}
\end{equation}
with conditional analogue
$\operatorname{TC}(\mathbf{H}_{1:N}\mid\theta)
:=\sum_i\mathcal{H}(H_i\mid\theta)-\mathcal{H}(\mathbf{H}_{1:N}\mid\theta)$.
For clarity, we suppress conditioning on the fixed policies in the identity below.

\begin{proposition}[Coordination gap]
\label{prop:coordination-gap}
For any joint distribution over $\theta$ and $\mathbf{H}_{1:N}$, assuming conditional independence
of the designer histories given $\theta$,
\begin{equation}
\sum_{i\in\mathcal{I}} I(\theta;H_i)
-
I(\theta;\mathbf{H}_{1:N})
=
\operatorname{TC}(\mathbf{H}_{1:N})
\geq 0.
\label{eq:coordination-gap}
\end{equation}
\end{proposition}

The proof is given in Appendix~\ref{app:proofs}. We use the quantity in Proposition~\ref{prop:coordination-gap} as a metric for \emph{coordination gap} in cooperative \framework\ with a shared inferential target: a diagnostic of the discrepancy between designer-wise and team-wise valuation of an experiment. This is not intended as the unique measure of coordination failure within \framework, but it is particularly natural for BED because it is expressed in the same information units as the underlying design objective. We report it alongside the information acquired jointly by the team.

Under Proposition~\ref{prop:coordination-gap}, independent optimisation over-counts information
by exactly the total correlation among designer histories. For two designers, the gap reduces to
$I(H_1;H_2)$. More generally,
$\operatorname{TC}(\mathbf{H}_{1:N})=\sum_i I(H_i;\mathbf{H}_{1:i-1})$, so the gap measures
the cumulative information in each designer's history that is predictable from the histories acquired
by the others. Although histories are independent once $\theta$ is known,
marginalising over $\theta$ can make them strongly dependent.

The coordination gap therefore captures a specific failure mode of cooperative \framework: designers may each
acquire highly informative data while collectively spending experimental capacity on overlapping
uncertainty. A small gap indicates close alignment between designer-wise and team-wise valuation,
whereas a large gap indicates substantial redundancy across the team.

\subsection{Heterogeneous Designers and Costs}
\label{subsec:heterogeneity}

The \framework\ formulation naturally accommodates designers with different design spaces, observation models, capabilities, constraints, and experimental costs. Here we consider the case where the costs are driven by the history and the design currently being made. Let $c_i(\xi_{i,t},H_{i,t-1})\geq0$ denote the cost incurred when designer $i$ selects design $\xi_{i,t}$. We consider a cost-aware objective (reminiscent of \citet{Hedman2026FairBEDAB})
\begin{equation}
\mathcal{J}_{\mathrm{cost}}(\boldsymbol{\pi})
:=
\operatorname{PEIG}(\boldsymbol{\pi})
-
\lambda\,
\mathbb{E}
\left[
\sum_{i\in\mathcal{I}}\sum_{t=1}^{T}
c_i(\xi_{i,t},H_{i,t-1})
\right],
\label{eq:cost-aware-objective}
\end{equation}
where $\lambda\geq0$ controls the information--cost trade-off.

Heterogeneity makes coordination an allocation problem as well as an information problem. The value of assigning an experiment to designer $i$ depends not only on that designer's own information--cost trade-off, but also on what the remaining designers are expected to learn and at what cost. A coordinated approach can therefore specialise designers according to their relative capabilities and costs rather than optimising each designer in isolation.

\section{Learning Coordinated Multi-Designer Policies}
\label{sec:learning}

Having identified the gap between designer-wise and team-wise information objectives, we now turn to learning policies that exploit the joint objective while preserving decentralised execution.

\subsection{CTDE for Multi-Designer BED}
\label{subsec:ctde}

Decentralised execution restricts designer \(i\) to decisions based on its own local history. The resulting coordination challenge is that the value of a designer's experiment depends on what the rest of the team is expected to acquire, even though those observations are unavailable at execution. Independent training does not account for this interaction, since each \(\pi_i\) is optimised only from the perspective of its own local information objective.

This suggests separating the information available when policies are \emph{trained} from the information available when they are \emph{executed}. We therefore adapt \emph{centralised training with decentralised execution} (CTDE) from the MARL literature to BED \citep{lowe2017multiagent,rashid2018qmix,hedman2026coda}. During training, we simulate the complete joint history $\mathbf{H}_{1:N}$ and optimise all constituent policies against the shared PEIG objective. At execution,~\eqref{eq:joint-policy} remains unchanged: designer $i$ selects designs using only $H_{i,t-1}$.

\looseness=-1
Centralisation therefore enters through the training signal rather than through execution-time communication. A local policy can learn that an experiment is useful because it complements what the rest of the team is expected to acquire, even though it never observes the other designers' histories at deployment. CTDE specifies this information structure rather than a particular architecture or optimiser; any factorised policy satisfying~\eqref{eq:joint-policy} can in principle be trained using a tractable estimator of PEIG.

\subsection{Multi-Designer Contrastive Estimation}
\label{subsec:contrastive}

Similarly to the single designer BED setting, direct optimisation of PEIG remains intractable because it requires evaluating
$p(\mathbf{H}_{1:N}\mid\boldsymbol{\pi})$.
We therefore extend the contrastive machinery of Section~\ref{sec:background_estimators} to the joint multi-designer history. We draw $\theta_0\sim p(\theta)$, simulate $\mathbf{H}_{1:N}\sim p(\mathbf{H}_{1:N}\mid\theta_0,\boldsymbol{\pi})$, and draw $L$ contrastive latents $\theta_{1:L}\overset{\mathrm{iid}}{\sim}p(\theta)$.
Subsequently, define the joint observation likelihood as in Eq.~\eqref{eq:md-history-factorisation}.
The resulting multi-designer prior-contrastive bound is
\begin{equation}
\mathcal{L}_{\mathrm{MD\text{-}PCE}}(\boldsymbol{\pi};L)
=
\mathbb{E}
\left[
\log
\frac{p(\mathbf{H}_{1:N}\mid\theta_0,\boldsymbol{\pi})}
{\frac{1}{L+1}\sum_{\ell=0}^{L}
p(\mathbf{H}_{1:N}\mid\theta_\ell,\boldsymbol{\pi})}
\right]
\leq
\operatorname{PEIG}(\boldsymbol{\pi}).
\label{eq:md-pce}
\end{equation}
When the simulator and policies are reparameterisable,~\eqref{eq:md-pce} can be differentiated through the complete simulated multi-designer acquisition process, yielding a single team-level training signal for all local policies.
 Note, independent training applies a separate single designer contrastive objective to each $H_i$, rewarding information regardless of whether another designer is expected to acquire the same evidence. MD-PCE instead scores $\mathbf{H}_{1:N}$ inside one contrastive objective. 

\subsection{DAD-Based Algorithms}
\label{subsec:algorithms}

We instantiate \framework\ with a family of DAD-based policies. Each designer maps its own
design--observation history through an encoder to a fixed-dimensional representation,
and an emitter maps that representation, optionally with designer attributes, to the
next design, so every variant satisfies the decentralised execution constraint
in~\eqref{eq:joint-policy}. Crossing the training objective with the parameterisation
gives four algorithms: \mddad\ and \mddadps\ optimise the joint MD-PCE objective
in~\eqref{eq:md-pce} with designer-specific and shared parameters respectively, while
\indad\ and \indadps\ are their independently trained counterparts, applying a
single-designer contrastive objective to each local history. Shared-parameter variants may
additionally receive a one-hot designer identifier, denoted by the -ID suffix, giving \mddadpsid\ and
\indadpsid, which allows one policy to distinguish the designer it is deployed as.
Table~\ref{tab:dad-variants} in Appendix~\ref{app:variants} summarises the family.

\section{Related Work}
\label{sec:rel_works}

\looseness=-1
\textbf{Bayesian experimental design.}
Recent BED has moved from optimising isolated designs toward amortised acquisition
policies. DAD learns a non-myopic policy over complete experimental trajectories
\citep{foster2021dad}, with subsequent work extending policy-based BED to implicit
likelihoods \citep{ivanova2021implicit}, semi-amortised adaptation
\citep{hedman2025stepdad}, reinforcement-learning formulations
\citep{blau2022optimizing,Shen2021BayesianSO}, joint amortisation of inference and
acquisition \citep{Huang2025ALINEJA}, task-driven objectives
\citep{rossa2026actionbedtaskdrivenbayesianexperimental, amortisedHuang}, and
fairness-aware acquisition \citep{Hedman2026FairBEDAB}. We are complementary to this
line: we ask how BED
should be formulated when experimental control is distributed across multiple designers. 

\looseness=-1
\textbf{Multi-actor information gathering and coordination.}
Decentralised information gathering, active sensing and multi-robot search provide the
closest precedent for coordinated acquisition without a central controller. Dec-POMDP
formulations study cooperative information gathering under local policies and limited
communication \citep{ma_info_gathering}, decentralised active search examines how actors
can diversify their measurements \citep{pmlr-v161-ghods21a}, and
\citet{Gupta-2026-88364} casts multi-robot active search as decentralised batch BED.
None develops distributed experimental control as a general BED perspective in which
locally adaptive histories are valued through a shared information objective. From MARL we inherit CTDE, which is standard
for navigating coordination under partial observability
\citep{lowe2017multiagent,rashid2018qmix,foerster2018coma,mappo,masac,matd3,Cooperative_MARL_deepRL},
and where coordination failure has been studied directly
\citep{hedman2026coda,tilbury2024coordinationfailurecooperativeoffline,MOMAPPO}. Those
methods optimise accumulated reward in a stochastic game; cooperative \framework\ instead optimises
a Bayesian information objective on the joint experiment, quantifies redundant
acquisition information-theoretically, and trains decentralised design policies through
the experimental generative model. In this work, execution requires no communication
\citep{Sukhbaatar2016LearningMC, multiagentlanguage}, though communication is a natural
extension of the \framework~paradigm.
\section{Experiments}
\label{sec:experiments}

We now evaluate \mddad\ and its broader CTDE framework. We study source localisation \citep{foster2021dad}, a standard benchmark for BED: $K=2$ hidden sources must be located from noisy observations. The total budget is held at $NT=12$ experiments for every method and every
team size, so $N$ designers make $T=12/N$ designs each. Further detail on the model, architectures and hyperparameters is in
Appendix~\ref{app:details}. Results are means and standard errors over three seeds.

\textbf{Metrics.}
We target PEIG \eqref{eq:peig} with MD-PCE \eqref{eq:md-pce} from below and MD-NMC \eqref{eq:md-nmc} from above; the lower bound is the optimisation objective. We also report the coordination gap of
Proposition~\ref{prop:coordination-gap}, $\sum_i I(\theta;H_i)-I(\theta;\mathbf{H}_{1:N})$.
Both of its terms are themselves targeted with bounds, so their difference is not itself a bound: we therefore
report the plug-in value $\sum_i \mathrm{sPCE}_i-\mathrm{sPCE}_{1:N}$ as a point
estimate, alongside the \emph{gap bracket}
$[\sum_i \mathrm{sPCE}_i-\mathrm{sNMC}_{1:N},\ \sum_i \mathrm{sNMC}_i-\mathrm{sPCE}_{1:N}]$,
the narrowest interval the same rollouts certify the true gap to lie in, up to Monte Carlo error.

\subsection{Coordination at a fixed team size}
\label{subsec:core}

We first test the central claim that centralising information during training produces more
coordinated acquisition.
Table~\ref{tab:core} shows that acquisition under policies trained to target PEIG (\mddad) increases the information in the acquired
dataset. \mddad\ scores higher than the independently optimised \indad, and the
same ordering holds under shared policy parameters (\mddadpsid\ vs \indadpsid). The
coordination gap is roughly halved by the joint objective, so the team gains more informative and less marginally duplicated designs.

\looseness=-1
Interestingly, this team gain arises while each designer becomes individually \emph{less}
informative. The per-designer EIGs in Table~\ref{tab:dad-budget} show that a jointly trained designer is individually less informative than its independently trained counterpart despite the team as a whole performing more strongly. This illustrates the coordination failure identified above. Note each independently trained designer is essentially a single-designer DAD policy given the same horizon. 
Despite their poor coordination and risk of duplicated designs, \indad\ and \indadpsid\ outcompete random design acquisition. 
Finally, a single designer spending the whole budget sequentially naturally performs most strongly. It faces
no multi-designer coordination problem and conditions each design on experimental history.
Its performance characterises the price of decentralisation. 

\begin{table}[t]
\caption{\textbf{Coordination at $N=3$.} PEIG bounds (mean $\pm$ s.e.\ over three
seeds). Single-designer DAD spends all twelve experiments itself, so it faces no
coordination problem and conditions each design on everything collected so far; it is
an upper reference rather than a direct comparison.}
\label{tab:core}
\centering\small
\begin{tabular}{lcccc}
\toprule
Method & PEIG (lower) & PEIG (upper) & Coordination gap & Gap bracket\\
\midrule
Random & $5.39 \pm 0.01$ & $5.40 \pm 0.01$ & n/a & n/a \\
\indadpsid & $6.63 \pm 0.03$ & $6.65 \pm 0.03$ & $4.79 \pm 0.03$ & $[4.78, 4.79]$ \\
\indad & $7.18 \pm 0.07$ & $7.24 \pm 0.07$ & $4.12 \pm 0.14$ & $[4.06, 4.12]$ \\
\mddadpsid & $8.43 \pm 0.02$ & $8.56 \pm 0.03$ & $2.11 \pm 0.02$ & $[1.97, 2.11]$ \\
\mddad & $8.45 \pm 0.02$ & $8.59 \pm 0.03$ & $2.07 \pm 0.04$ & $[1.93, 2.07]$ \\
\midrule
DAD (one designer, $T=12$) & $9.81 \pm 0.06$ & $10.27 \pm 0.12$ & n/a & n/a \\
\bottomrule
\end{tabular}
\end{table}

\subsection{Scaling with team size}
\label{subsec:scaling}

\begin{figure}[t]
\centering
\includegraphics[width=0.88\linewidth]{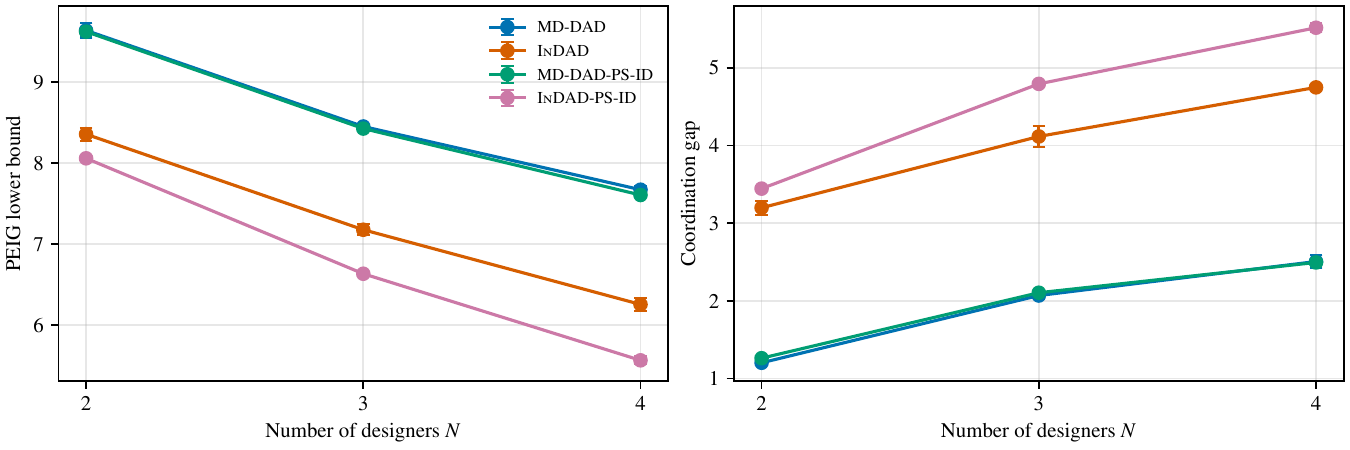}
\caption{\textbf{Coordination as the team grows.} PEIG lower bound (left) and
coordination gap (right) against team size at a fixed budget of $12$ experiments, so
additional designers take correspondingly fewer rounds each. Error bars are one standard
error over three seeds; Table~\ref{tab:scaling} for per-point values.}
\label{fig:scaling}
\end{figure}

Coordination is hardest when there are more designers to coordinate, so we next ask how
robust the effect is as the team grows, with the budget fixed at $NT=12$ so that adding
designers redistributes data rather than adding it. Figure~\ref{fig:scaling} shows the
benefit of centralised training holds at every team size, with the parameter-sharing
variants following the same pattern. Every method loses performance as designers are added
and every coordination gap widens, but the margin between the objectives widens with it:
the harder coordination becomes, the more the joint objective is worth. Whether the
decline itself reflects harder coordination or weaker adaptivity is separable, and we
ablate it in Appendix~\ref{app:dad-budget}.

\subsection{Designer identifiers under parameter sharing}
\label{subsec:id}

\begin{figure}[t]
\centering
\includegraphics[width=0.88\linewidth]{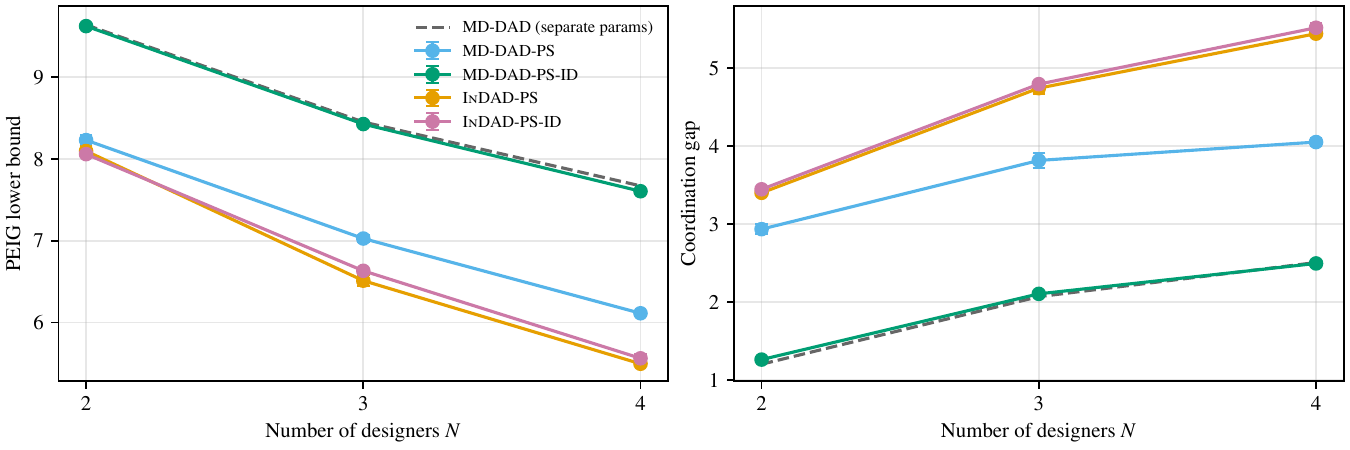}
\caption{\textbf{Designer identifiers under parameter sharing.} Shared-parameter variants
with and without the one-hot identifier against team size, with separately parameterised
\mddad\ dashed for reference. Panels and error bars follow Figure~\ref{fig:scaling};
per-point values are in Table~\ref{tab:id}.}
\label{fig:id}
\end{figure}

Our designers experience homogeneous settings, so the parameter-sharing variants of
Section~\ref{subsec:algorithms} cannot natively tell themselves apart. A natural remedy is to give the shared
policy a one-hot designer identifier, giving the \textsc{-ID} variants
(Appendix~\ref{app:variants}). We explore the effect of this on designer specialisation.
Figure~\ref{fig:id} shows that under the
joint objective the identifier closes almost all of the shortfall against separate
parameters at every team size; under the independent objective the same architectural
change has no effect. This is what the mechanism predicts: an identifier supplies the
\emph{capacity} to differentiate, while the \emph{incentive} must come from the
objective. Designers rewarded only for their own information have no reason to occupy
distinct roles.

\subsection{Heterogeneous costs}
\label{subsec:cost}

\begin{table}[t]
\caption{\textbf{Heterogeneous costs} with two designers, each objective with the penalty
off and on; the $\lambda=0$ rows measure cost without penalising it and are the
controls. Separation is the distance between the two designers' mean measurement locations
within a rollout, a geometric proxy for how far they divide the design space between
them, and spread is the mean distance of a designer's measurements from its own centre.
Per-designer costs and the measurements themselves are in
Appendix~\ref{app:cost-designs}. Figure~\ref{fig:cost-designs} provides an intuitive visualisation.}
\label{tab:cost}
\centering\small
\setlength{\tabcolsep}{2.5pt}
\begin{tabular}{lcccccc}
\toprule
& & & & & \multicolumn{2}{c}{Spread}\\
\cmidrule(lr){6-7}
Method & PEIG (lower) & Total cost & Coord.\ gap & Separation & Cheap & Expensive\\
\midrule
\indad\ ($\lambda=0$)   & $8.34 \pm 0.07$ & $6.83 \pm 0.10$ & $3.11 \pm 0.07$ & $0.52 \pm 0.03$ & $0.55 \pm 0.03$ & $0.53 \pm 0.03$ \\
\indad\ ($\lambda=0.1$) & $8.35 \pm 0.15$ & $6.20 \pm 0.21$ & $3.08 \pm 0.11$ & $0.52 \pm 0.03$ & $0.54 \pm 0.01$ & $0.51 \pm 0.02$ \\
\midrule
\mddad\ ($\lambda=0$)   & $9.56 \pm 0.07$ & $8.34 \pm 0.15$ & $1.25 \pm 0.03$ & $1.33 \pm 0.04$ & $0.64 \pm 0.02$ & $0.70 \pm 0.02$ \\
\mddad\ ($\lambda=0.1$) & $9.55 \pm 0.02$ & $6.49 \pm 0.02$ & $1.32 \pm 0.03$ & $1.23 \pm 0.04$ & $0.69 \pm 0.03$ & $0.61 \pm 0.02$ \\
\bottomrule
\end{tabular}
\end{table}

Finally we ask whether centralisation can enable coordinated \emph{designer specialisation} and the
allocation of roles by capability. To test this we let designers differ in what an
experiment costs them (Eq.~\eqref{eq:cost-aware-objective}), with two designers traversing
the design space at travel-cost rates differing ninefold
(Appendix~\ref{app:cost-designs}). Under these constraints, we test whether a team learns to leave
wide-ranging exploration to the cheaper designer, and confine the movement of the expensive designer, all while maintaining performance.

Table~\ref{tab:cost} shows that both objectives cut spending and pay for it negligibly
in information. Only the jointly trained team, however, adapts its strategy to the
distinct capabilities of its designers, a strong sign of coordination. Note first that
\mddad's larger separation, the distance between the designers' mean measurement locations,
is a demonstration of coordination even before any cost signal: the designers carve up the
design space so that each works with minimal overlap
(Figure~\ref{fig:cost-designs}). Under cost pressure ($\lambda: 0 \to 0.1$) there is then
further specialisation. The cheap designer's spread increases while the expensive designer's
decreases; demonstrating designer specialisation. The larger exploratory
trajectories are handed to the cheaper designer and the expensive designer confined to a small
cluster of locally informative designs (Figure~\ref{fig:cost-designs} shows this clearly
in the design space). By contrast, the independently optimised designers minimise cost by
reducing exploration for both policies uniformly, with spread falling in both cases,
never deciding who should explore. Specialisation is therefore
itself a form of coordination rather than merely a response to the penalty, and \framework\
supplies the objective under which it is learned.
\section{Conclusion}
\label{sec:conclusion}

In this work we introduce \framework, a paradigm for Bayesian experimental design in
which experimental control is distributed across multiple designers. Within this paradigm we focus on the cooperative, decentralised setting where data is  \emph{acquired locally but valued jointly} for a shared inferential target. We show that
optimising each designer against its own information overvalues acquisition by exactly the
predictive total correlation among designer histories, so coordination joins as a new axis along which experimental design can struggle. Our algorithms navigate coordination failure by
separating what a policy may use when trained from what it uses when it acts (centralised training with decentralised execution), optimising
a contrastive bound on the introduced PEIG objective while each designer still designs from its
own history alone. Jointly trained teams consequently acquire substantially more pooled
information than independently trained policies. Under heterogeneous costs they learn further \emph{designer
specialisation}, confining the designer for which experiments are expensive to a smaller
design region while its cheaper partner explores, where independently trained
designers facing the same costs merely economise uniformly.

\begin{ack}
MH is supported by funding provided by Novo Nordisk and by the EPSRC Centre for Doctoral Training in Modern Statistics and Statistical Machine Learning.
\end{ack}

\bibliographystyle{plainnat}
\bibliography{iclr2027_conference}


\newpage
\appendix
\section{Additional related work}
\label{app:related}

Section~\ref{sec:rel_works} concentrates on the work \framework\ builds on directly. For
completeness we record here additional BED literature that our work inherits
from but does not modify.

\paragraph{Bayesian experimental design.}
BED selects experiments by maximising an expected utility, most commonly the expected
information gain \citep{lindley1956,chaloner1995,ryan2016review,rainforth2024modern}. 
The expected information gain is doubly intractable, and a substantial computational
literature addresses it: nested Monte Carlo and its variants
\citep{pmlr-v80-rainforth18a,iqbal2024nesting}, variational and stochastic-gradient
methods \citep{foster2019variational,foster2020unified}, contrastive bounds of the kind
used throughout this paper \citep{foster2020unified,foster2021dad}, likelihood-free and
mutual-information-estimation approaches
\citep{kleinegesse2019efficient,Kleinegesse2020BayesianED}, and particle-based
sequential schemes \citep{iollo2024pasoa}. \framework\ is agnostic to this choice. We build on top of
sPCE and sNMC because they are heavily employed in recent work. Any tractable estimator of PEIG could be
substituted.

\section{Proofs}
\label{app:proofs}

\subsection{Proof of Proposition~\ref{prop:coordination-gap}}

\begin{proof}
Write each mutual information in entropy form, $I(\theta;X)=\mathcal H(X)-\mathcal H(X\mid\theta)$,
and regroup the marginal and conditional terms:
\begin{align}
\sum_{i \in \mathcal{I}} I(\theta; H_i) - I(\theta; \mathbf{H}_{1:N})
&= \sum_{i \in \mathcal{I}}\left[\mathcal{H}(H_i) - \mathcal{H}(H_i \mid \theta)\right]
 - \left[\mathcal{H}(\mathbf{H}_{1:N}) - \mathcal{H}(\mathbf{H}_{1:N} \mid \theta)\right]
\nonumber\\
&= \left[\sum_{i \in \mathcal{I}} \mathcal{H}(H_i) - \mathcal{H}(\mathbf{H}_{1:N})\right]
 - \left[\sum_{i \in \mathcal{I}} \mathcal{H}(H_i \mid \theta) - \mathcal{H}(\mathbf{H}_{1:N} \mid \theta)\right]
\nonumber\\
&= \operatorname{TC}(\mathbf{H}_{1:N}) - \operatorname{TC}(\mathbf{H}_{1:N} \mid \theta),
\label{eq:general-gap}
\end{align}
by the definition in Eq.~\eqref{eq:total-correlation} and its conditional analogue. This
identity holds for any joint distribution over $\theta$ and $\mathbf{H}_{1:N}$.

It remains to show that the second term vanishes under the assumption of the
proposition. Both total correlations are KL divergences,
\begin{equation}
\operatorname{TC}(\mathbf{H}_{1:N} \mid \theta)
=
\mathbb{E}_{p(\theta)}
\left[
\operatorname{KL}\!\left(
p(\mathbf{H}_{1:N} \mid \theta)
\,\middle\|\,
\prod_{i \in \mathcal{I}} p(H_i \mid \theta)
\right)
\right],
\label{eq:conditional-tc-kl}
\end{equation}
and likewise for $\operatorname{TC}(\mathbf{H}_{1:N})$ with the marginals in place of the
conditionals. Conditional independence of the designer histories given $\theta$ states
exactly that $p(\mathbf{H}_{1:N} \mid \theta) = \prod_{i} p(H_i \mid \theta)$, so the
divergence in Eq.~\eqref{eq:conditional-tc-kl} is zero for $p(\theta)$-almost every
$\theta$ and hence $\operatorname{TC}(\mathbf{H}_{1:N} \mid \theta) = 0$. Substituting
into Eq.~\eqref{eq:general-gap} gives
\begin{equation}
\sum_{i \in \mathcal{I}} I(\theta; H_i) - I(\theta; \mathbf{H}_{1:N})
= \operatorname{TC}(\mathbf{H}_{1:N}),
\nonumber
\end{equation}
which is the claimed identity, and $\operatorname{TC}(\mathbf{H}_{1:N}) \geq 0$ because it
is a KL divergence, giving Eq.~\eqref{eq:coordination-gap}.
\end{proof}

In our setting the required conditional independence of local histories given $\theta$ is
not an extra assumption but a consequence of the acquisition structure: since each designer's outcomes
depend only on its own history,
Eq.~\eqref{eq:md-history-factorisation} factorises
$p(\mathbf{H}_{1:N} \mid \theta,\boldsymbol{\pi})$ over designers. Eq.~\eqref{eq:general-gap} itself needs no such
assumption, so the coordination gap remains an exact characterisation for conditionally dependent local histories: it is
then the signed difference of two total correlations rather than a measure of redundancy
alone.

\section{Replicated parameter sharing policies are a subset of decentralised teams}
\label{app:replication}

Let $\Pi$ be the class of local policies, $\Pi_{\mathrm{dec}} = \Pi^N$ the decentralised
teams whose designers may use different local policies, and
$
\Pi_{\mathrm{rep}} = \{(\pi,\ldots,\pi) : \pi \in \Pi\}
$
the subclass in which the same policy is deployed by every designer, as under parameter
sharing without a designer identifier.

\begin{corollary}

$$
\sup_{\boldsymbol{\pi} \in \Pi_{\mathrm{dec}}}
\operatorname{PEIG}(\boldsymbol{\pi})
\geq
\sup_{\boldsymbol{\pi} \in \Pi_{\mathrm{rep}}}
\operatorname{PEIG}(\boldsymbol{\pi}).
$$

\end{corollary}

\begin{proof}
Every replicated tuple $(\pi,\ldots,\pi)$ lies in $\Pi^N$, so
$\Pi_{\mathrm{rep}} \subseteq \Pi_{\mathrm{dec}}$. Therefore, the supremum of the same
objective over $\Pi_{\mathrm{dec}}$ cannot be smaller than the supremum over its
subset $\Pi_{\mathrm{rep}}$.
\end{proof}

This result concerns representational capacity rather than optimisation or sample
efficiency. Since $\Pi_{\mathrm{rep}} \subseteq \Pi_{\mathrm{dec}}$, restricting all
designers to a shared policy cannot increase the best achievable PEIG. It therefore does
not by itself predict the empirical ordering of \mddadps\ and \mddad\ in
Section~\ref{subsec:id}.

In practice, parameter sharing can nevertheless be advantageous because experience
collected by different designers is used to update the same policy, improving sample
efficiency and reducing the number of parameters that must be learned. The comparison
therefore reflects a trade-off: fully decentralised policies provide a larger
representable policy class, whereas parameter sharing imposes a useful restriction
that can make learning more data-efficient. Supplying a designer identifier allows a
shared policy to condition its behaviour on designer identity, recovering designer-specific
behaviour while retaining shared parameters.

\section{Multi-Designer BED Sub-Settings}
\label{app:mdbed-settings}

Table~\ref{tab:mdbed-settings} summarises select structural dimensions
along which a \framework~problem may vary. A particular
application is specified by selecting one regime along each axis.

\arrayrulecolor{black!35}

\begin{table*}[t]
\centering
\caption{
\textbf{A taxonomy of multi-designer Bayesian experimental design (\framework) settings.}
Each row defines an approximately orthogonal structural dimension of an \framework\ problem.
The setting studied in this work is shown in \textbf{bold}.
}
\label{tab:mdbed-settings}
\small
\renewcommand{\arraystretch}{1.12}
\setlength{\tabcolsep}{4pt}

\begin{tabularx}{\textwidth}{
    >{\raggedright\arraybackslash}p{0.17\textwidth}
    >{\raggedright\arraybackslash}p{0.25\textwidth}
    >{\raggedright\arraybackslash}X
}
\arrayrulecolor{black}
\toprule
\textbf{Dimension} & \textbf{Possible regimes} & \textbf{Interpretation} \\
\midrule
\arrayrulecolor{black!35}
Objective structure
&
\parbox[t]{\linewidth}{\raggedright
\textbf{Fully cooperative}\\
Mixed-motive\\
Competitive
}
&
Whether designers optimise a common inferential objective, partially aligned objectives, or opposing objectives. In the cooperative setting studied here, all acquired data contribute to the same team-level inferential utility.
\\
\specialrule{0.3pt}{0pt}{0pt}

Execution information $\mathcal{F}_{i,t-1}$
&
\parbox[t]{\linewidth}{\raggedright
Centralised\\
\textbf{Decentralised}\\
Partially shared
}
&
Whether experimental decisions may condition on the complete joint history, only a designer's local history, or an intermediate information set containing selected information from other designers.
\\
\specialrule{0.3pt}{0pt}{0pt}

Communication
&
\parbox[t]{\linewidth}{\raggedright
\textbf{None}\\
Intermittent\\
Continuous
}
&
Whether designers may exchange observations, posterior summaries, intended designs, or other messages during acquisition.
\\
\specialrule{0.3pt}{0pt}{0pt}

Interaction timing
&
\parbox[t]{\linewidth}{\raggedright
\textbf{Synchronous}\\
Asynchronous\\
Event-driven
}
&
Whether designers select experiments in coordinated rounds, operate on independent clocks, or acquire data when designer- or environment-specific events permit.
\\
\specialrule{0.3pt}{0pt}{0pt}

Inferential target
&
\parbox[t]{\linewidth}{\raggedright
\textbf{Shared latent target}\\
Partially shared targets\\
Designer-specific targets
}
&
Whether designers seek information about the same latent quantity, overlapping components of a larger latent state, or distinct quantities coupled through a common model.
\\
\specialrule{0.3pt}{0pt}{0pt}

Acquisition dynamics
&
\parbox[t]{\linewidth}{\raggedright
History-independent\\
\textbf{Local-history dependent}\\
Cross-designer-history dependent
}
&
Whether an observation depends only on the current design and latent quantity, additionally on the acting designer's previous experiments, or on histories and interventions across multiple designers.
\\
\specialrule{0.3pt}{0pt}{0pt}

Designer capabilities
&
\parbox[t]{\linewidth}{\raggedright
Homogeneous\\
\textbf{Heterogeneous}
}
&
Whether designers share the same design space and observation process or differ in sensing precision, experimental actions, noise models, constraints, or other acquisition capabilities.
\\
\specialrule{0.3pt}{0pt}{0pt}

Policy parameterisation
&
\parbox[t]{\linewidth}{\raggedright
\textbf{Designer-specific}\\
\textbf{Shared}\\
\textbf{Shared + designer-conditioned}
}
&
Whether designers use separate policy parameters, a common policy, or a shared policy conditioned on designer attributes or identities.
\\
\specialrule{0.3pt}{0pt}{0pt}

Resource structure
&
\parbox[t]{\linewidth}{\raggedright
Independent budgets\\
Shared team budget\\
\textbf{Designer-specific costs}\\
Mixed constraints
}
&
Whether experimental resources are allocated independently, drawn from a common team budget, priced differently across designers, or constrained through combinations of local and global resource limits.
\\
\specialrule{0.3pt}{0pt}{0pt}

Team composition
&
\parbox[t]{\linewidth}{\raggedright
\textbf{Fixed}\\
Variable\\
Open
}
&
Whether the identities and number of designers are known throughout acquisition, vary between experimental instances, or may change during the experiment through arrivals, failures, or departures.
\\
\arrayrulecolor{black}
\bottomrule
\end{tabularx}
\end{table*}

\arrayrulecolor{black}

\section{Estimating the multi-designer objective}
\label{app:estimators}

This section states the estimators used to train and evaluate every method.

\subsection{MD-PCE and MD-NMC}
\label{app:md-pce-nmc}

Let $\theta_0 \sim p(\theta)$ generate a joint history $\mathbf{H}_{1:N}$ under the factorised
policy $\boldsymbol\pi$, and let $\theta_{1:L}\overset{\mathrm{iid}}\sim p(\theta)$ be
contrastive latents. With the joint observation likelihood
$p(\mathbf{H}_{1:N}\mid\theta,\boldsymbol{\pi})$ of Eq.~\eqref{eq:md-history-factorisation}, the
multi-designer prior-contrastive bound of Eq.~\eqref{eq:md-pce} is
\begin{equation}
\mathcal{L}_{\mathrm{MD\text{-}PCE}}(\boldsymbol{\pi};L)
=
\mathbb{E}
\left[
\log
\frac{p(\mathbf{H}_{1:N}\mid\theta_0,\boldsymbol{\pi})}
{\frac{1}{L+1}\sum_{\ell=0}^{L}
p(\mathbf{H}_{1:N}\mid\theta_\ell,\boldsymbol{\pi})}
\right]
\leq
\operatorname{PEIG}(\boldsymbol{\pi}).
\end{equation}
and the corresponding nested Monte Carlo estimator excludes the positive sample from
the marginal average,
\begin{equation}
\mathcal{U}_{\mathrm{MD\text{-}NMC}}(\boldsymbol{\pi};L)
=
\mathbb{E}
\left[
\log
\frac{p(\mathbf{H}_{1:N}\mid\theta_0,\boldsymbol{\pi})}
{\frac{1}{L}\sum_{\ell=1}^{L}
p(\mathbf{H}_{1:N}\mid\theta_\ell,\boldsymbol{\pi})}
\right]
\geq
\operatorname{PEIG}(\boldsymbol{\pi}).
\label{eq:md-nmc}
\end{equation}
 Both are computed from
one set of contrastive draws, since they differ only in whether $\ell=0$ enters the
denominator, and together they bracket PEIG. MD-PCE is bounded above by
$\log(L+1)$, so a bound approaching that ceiling reports the contrastive budget rather
than the policy; we monitor this saturation directly.

The independent objective, which scores each designer against the information in its
own local history alone, is estimated by applying the same PCE bound to each designer's
local history and summing,
\begin{equation}
\mathcal{L}_{\mathrm{ind}}(\boldsymbol{\pi};L)
=
\sum_{i \in \mathcal{I}}
\mathbb{E}
\left[
\log
\frac{p(H_i \mid \theta_0, \pi_i)}
{\frac{1}{L+1}\sum_{\ell=0}^{L}
p(H_i \mid \theta_\ell, \pi_i)}
\right].
\label{eq:ind-pce}
\end{equation}
During training each designer's term draws its own contrastive latents, which keeps the
$N$ terms independent. At evaluation the joint and per-designer bounds share one set of
draws, since there the two are differenced to form the coordination gap and pairing
them removes Monte Carlo variance from that difference.

\section{DAD-based variants}
\label{app:variants}

Table~\ref{tab:dad-variants} summarises the family of algorithms introduced in
Section~\ref{subsec:algorithms}. The two axes are orthogonal: the training objective
determines whether a designer is rewarded for information the rest of the team is also
expected to acquire, and the parameterisation determines whether designers can differ.
Shared-parameter variants may additionally take a one-hot designer identifier, denoted by the
\textsc{-ID} suffix, which is what allows one set of weights to act as distinguishable
designers.

\begin{table}[t]
\centering
\caption{\textbf{DAD-based variants for \framework.}}
\label{tab:dad-variants}
\small
\begin{tabular}{lcc}
\toprule
& \textbf{Separate parameters} & \textbf{Shared parameters} \\
\midrule
\makecell[l]{\textbf{Joint objective}\\(MD-PCE)}
& \mddad
& \mddadps \\
\arrayrulecolor{black!35}
\midrule
\arrayrulecolor{black}
\makecell[l]{\textbf{Per-designer objective}\\(Single designer PCE)}
& \indad
& \indadps \\
\bottomrule
\end{tabular}
\end{table}

\section{Experimental details}
\label{app:details}
\paragraph{Implementation.}
Our implementation of \framework\ is built on top of TestBED \citep{testbed}, an open-source
toolkit for Bayesian experimental design, which we extend with multi-designer policies,
objectives and estimators.

\paragraph{Optimisation.}
Policies are trained with Adam under the settings in Table~\ref{tab:hparams}, using the
architecture in Tables~\ref{tab:arch-encoder} and~\ref{tab:arch-emitter}. Both are held
fixed across every method and team size, so comparisons hold the policy class and the
simulator budget fixed. 

\paragraph{Evaluation.}
Evaluation uses fresh prior and observation samples drawn after training, with
$2{,}048$ outer samples and $10^6$ contrastive samples per bound. MD-PCE and
MD-NMC provide the lower and upper bound. Reported uncertainty is the standard error across independently trained
seeds, not across Monte Carlo batches from a single checkpoint. Standard errors for
the coordination gap are computed from per-sample differences, since every term is
scored on the same rollout under the same latent draw and the difference is therefore
strongly paired.

\paragraph{Source localisation.}
A measurement at location $\xi \in \mathbb{R}^p$ returns
\begin{equation}
 y\sim\mathcal N\!\left(
 \log\left(b+\sum_{k=1}^K(\epsilon+\|\xi-\theta_k\|_2^2)^{-1}\right),
 \sigma^2\right),                                           \label{eq:location}
\end{equation}
the total intensity of $K$ sources at unknown locations $\theta_k$, observed on a log
scale in noise. We use $K=2$ sources and $p=2$, so each design $\xi$ and each source location $\theta_k$ lies in $\mathbb{R}^2$, with prior $\theta_k\sim\mathcal N(0,I)$, $b=0.1$,
$\epsilon=10^{-4}$, and $\sigma=0.5$ in Eq.~\eqref{eq:location}. The total budget is
fixed at $NT=12$ measurements, giving $T=6$ for $N=2$, $T=4$ for $N=3$ and $T=3$ for
$N=4$. 

\paragraph{Compute.}
Every run trains one team on a single GPU and is in the order of magnitude of hours. 

\paragraph{Policy architecture.}
The original DAD encoder pools a history with a DeepSets-style sum \citep{foster2021dad, zaheer2017deep},
making the representation permutation invariant: a history is a set of design--outcome
pairs and the posterior it induces does not depend on the order they were collected in.
Later policies replace that sum with attention over the same pairs
\citep{ivanova2021implicit}, which keeps the posterior-sufficiency intuition while
letting the pairs inform one another rather than being embedded in isolation. We follow
that line and add a learned embedding of the round index to every token, including the
readout token, which carries the current history length. Order invariance is therefore
given up deliberately, in exchange for the policy knowing how far through the experiment
it is. At a fixed budget the rounds already spent determine the rounds remaining, and
the optimal design depends on that: a designer with one round left should place its final
measurement to resolve as much residual uncertainty as it can, whereas the same designer at
the start can afford a design that is uninformative in itself but positions it to
exploit what it learns. A permutation-invariant encoder can condition on the difference
only indirectly, through the size of the history handed to it. The distinction matters
more here than for a single designer, since a fixed total budget divided among $N$ designers
leaves each of them fewer rounds over which to plan.

\paragraph{Hyperparameters and architecture.}
Table~\ref{tab:hparams} gives the training hyperparameters and
Tables~\ref{tab:arch-encoder} and~\ref{tab:arch-emitter} the architecture, all shared
by every method and team size. Every designer holds one encoder--emitter pair under
separate parameterisation, and the team holds one such pair under parameter sharing.

\begin{table}[t]
\caption{\textbf{Training hyperparameters.} Shared by every team size.}
\label{tab:hparams}
\centering\small
\begin{tabular}{lc}
\toprule
Parameter & Source localisation \\
\midrule
Contrastive samples, $L$ (training) & $65{,}536$ \\
Outer samples (training)            & $1{,}024$  \\
Initial learning rate               & $10^{-4}$  \\
Betas                               & $(0.9, 0.999)$ \\
Learning-rate schedule              & none \\
Gradient-norm clip                  & $1.0$ \\
Gradient steps                      & $50{,}000$ \\
Total budget, $NT$                  & $12$ \\
Designers, $N$                         & $\{2,3,4\}$\\
Rounds per designer, $T$               & $\{6,4,3\}$\\
Contrastive samples (evaluation)    & $10^{6}$ \\
Outer samples (evaluation)          & $2{,}048$ \\
Seeds                               & $3$ \\
\bottomrule
\end{tabular}
\end{table}

\begin{table}[t]
\caption{\textbf{Encoder $E_{\phi_1}$.} An attention-based history encoder applied to
designer $i$'s own local history. Each round of that history contributes one token, and a
learned embedding of the round index is added to it. The readout is a learned
\textsc{cls} token carrying the embedding of the current history length, so the policy
knows how many of its rounds remain; its output after the final block is the history
representation $R(H_{i,t-1})$. An empty history is represented by a learned null
embedding.}
\label{tab:arch-encoder}
\centering\small
\begin{tabular}{llcc}
\toprule
Layer & Description & Dimension & Activation\\
\midrule
Input & $(\xi_{i,t}, y_{i,t})$ per round & $3$ & --\\
H1 & Fully connected & $256$ & GELU\\
Token & Fully connected & $128$ & --\\
Time embedding & Learned, added to each token & $128$ & --\\
\textsc{cls} & Learned query token & $128$ & --\\
Blocks & $3\times$ pre-norm Transformer, $4$ heads & $128$ & GELU\\
Feed-forward & Within each block & $512$ & GELU\\
Output & LayerNorm on the \textsc{cls} output & $128$ & --\\
\bottomrule
\end{tabular}
\end{table}

\begin{table}[t]
\caption{\textbf{Emitter $F_{\phi_2}$.} Maps the history representation to the next
design. Under the \textsc{-ps-id} variants a one-hot designer identifier $z_i$ is
concatenated to the input, widening it to $128+N$; the first linear layer is then
itself the per-designer embedding table. }
\label{tab:arch-emitter}
\centering\small
\begin{tabular}{llcc}
\toprule
Layer & Description & Dimension & Activation\\
\midrule
Input & $R(H_{i,t-1})$, optionally with $z_i$ & $128$ $(+N)$ & --\\
H1 & Fully connected & $128$ & GELU\\
Output & $\xi_{i,t}$ & $2$ & identity\\
\bottomrule
\end{tabular}
\end{table}

\section{Per-point results}
\label{app:tables}

Tables~\ref{tab:scaling} and~\ref{tab:id} give the values plotted in
Figures~\ref{fig:scaling} and~\ref{fig:id}.

\begin{table}[t]
\caption{\textbf{Scaling with team size.} PEIG (mean $\pm$ s.e.\ over
three seeds) at a fixed total budget of $12$ measurements.}
\label{tab:scaling}
\centering\small
\begin{tabular}{llcccc}
\toprule
$N$ & Method & PEIG (lower) & PEIG (upper) & Coordination gap & Gap bracket\\
\midrule
2 & \mddad & $9.64 \pm 0.09$ & $10.55 \pm 0.12$ & $1.20 \pm 0.02$ & $[0.29, 1.21]$ \\
3 & \mddad & $8.45 \pm 0.02$ & $8.59 \pm 0.03$ & $2.07 \pm 0.04$ & $[1.93, 2.07]$ \\
4 & \mddad & $7.67 \pm 0.05$ & $7.71 \pm 0.05$ & $2.51 \pm 0.08$ & $[2.46, 2.51]$ \\
\midrule
2 & \indad & $8.36 \pm 0.08$ & $8.58 \pm 0.12$ & $3.20 \pm 0.09$ & $[2.98, 3.20]$ \\
3 & \indad & $7.18 \pm 0.07$ & $7.24 \pm 0.07$ & $4.12 \pm 0.14$ & $[4.06, 4.12]$ \\
4 & \indad & $6.25 \pm 0.08$ & $6.28 \pm 0.08$ & $4.75 \pm 0.05$ & $[4.73, 4.75]$ \\
\midrule
2 & \mddadpsid & $9.62 \pm 0.03$ & $10.41 \pm 0.04$ & $1.26 \pm 0.02$ & $[0.47, 1.27]$ \\
3 & \mddadpsid & $8.43 \pm 0.02$ & $8.56 \pm 0.03$ & $2.11 \pm 0.02$ & $[1.97, 2.11]$ \\
4 & \mddadpsid & $7.61 \pm 0.02$ & $7.65 \pm 0.02$ & $2.50 \pm 0.05$ & $[2.45, 2.50]$ \\
\midrule
2 & \indadpsid & $8.06 \pm 0.00$ & $8.14 \pm 0.02$ & $3.45 \pm 0.02$ & $[3.36, 3.45]$ \\
3 & \indadpsid & $6.63 \pm 0.03$ & $6.65 \pm 0.03$ & $4.79 \pm 0.03$ & $[4.78, 4.79]$ \\
4 & \indadpsid & $5.56 \pm 0.05$ & $5.57 \pm 0.05$ & $5.52 \pm 0.06$ & $[5.51, 5.52]$ \\
\bottomrule
\end{tabular}
\end{table}

\begin{table}[t]
\caption{\textbf{Designer identifiers under parameter sharing.} PEIG
(mean $\pm$ s.e.\ over three seeds). The separate-parameter \mddad\ rows are the
reference.}
\label{tab:id}
\centering\small
\begin{tabular}{llcccc}
\toprule
$N$ & Method & PEIG (lower) & PEIG (upper) & Coordination gap & Gap bracket\\
\midrule
2 & \mddadps & $8.23 \pm 0.06$ & $8.42 \pm 0.08$ & $2.93 \pm 0.07$ & $[2.75, 2.94]$ \\
3 & \mddadps & $7.03 \pm 0.04$ & $7.06 \pm 0.04$ & $3.82 \pm 0.10$ & $[3.78, 3.82]$ \\
4 & \mddadps & $6.11 \pm 0.03$ & $6.12 \pm 0.03$ & $4.05 \pm 0.05$ & $[4.04, 4.05]$ \\
\midrule
2 & \mddadpsid & $9.62 \pm 0.03$ & $10.41 \pm 0.04$ & $1.26 \pm 0.02$ & $[0.47, 1.27]$ \\
3 & \mddadpsid & $8.43 \pm 0.02$ & $8.56 \pm 0.03$ & $2.11 \pm 0.02$ & $[1.97, 2.11]$ \\
4 & \mddadpsid & $7.61 \pm 0.02$ & $7.65 \pm 0.02$ & $2.50 \pm 0.05$ & $[2.45, 2.50]$ \\
\midrule
2 & \indadps & $8.10 \pm 0.06$ & $8.20 \pm 0.07$ & $3.40 \pm 0.02$ & $[3.30, 3.41]$ \\
3 & \indadps & $6.51 \pm 0.07$ & $6.52 \pm 0.07$ & $4.74 \pm 0.07$ & $[4.74, 4.74]$ \\
4 & \indadps & $5.50 \pm 0.01$ & $5.50 \pm 0.00$ & $5.44 \pm 0.04$ & $[5.44, 5.44]$ \\
\midrule
2 & \indadpsid & $8.06 \pm 0.00$ & $8.14 \pm 0.02$ & $3.45 \pm 0.02$ & $[3.36, 3.45]$ \\
3 & \indadpsid & $6.63 \pm 0.03$ & $6.65 \pm 0.03$ & $4.79 \pm 0.03$ & $[4.78, 4.79]$ \\
4 & \indadpsid & $5.56 \pm 0.05$ & $5.57 \pm 0.05$ & $5.52 \pm 0.06$ & $[5.51, 5.52]$ \\
\midrule
2 & \mddad & $9.64 \pm 0.09$ & $10.55 \pm 0.12$ & $1.20 \pm 0.02$ & $[0.29, 1.21]$ \\
3 & \mddad & $8.45 \pm 0.02$ & $8.59 \pm 0.03$ & $2.07 \pm 0.04$ & $[1.93, 2.07]$ \\
4 & \mddad & $7.67 \pm 0.05$ & $7.71 \pm 0.05$ & $2.51 \pm 0.08$ & $[2.46, 2.51]$ \\
\bottomrule
\end{tabular}
\end{table}

\section{Where the designers measure under heterogeneous costs}
\label{app:cost-designs}

\paragraph{Motivation.}
In practice, designer coordination can extend beyond just avoiding redundancy (i.e. designers covering different parts of
the design space because information acquired twice is counted once). Heterogeneous costs
pose a second and distinct problem, of allocating roles to the designers best able to
perform them. If one designer can move cheaply and another cannot, the efficient division
is for the cheap designer to explore widely while the expensive one exploits locally, taking
a small number of measurements that are informative without requiring it to relocate.
Nothing in the objective states that division; it would have to be discovered from the
cost structure alone. This experiment explores this (initially discussed in Section~\ref{subsec:cost}).

\paragraph{Setting.}
Two designers share the source-localisation task of Section~\ref{sec:experiments} and the
same twelve-experiment budget, six rounds each, and differ only in what movement costs
them. Each pays $c_i = \rho_i \lVert \xi_{i,t} - \xi_{i,t-1} \rVert$ to relocate between
consecutive measurements, with rates $\rho_1 = 0.2$ for the inexpensive designer and
$\rho_2 = 1.8$ for the expensive one. The rates average to one, so $\lambda$ in
Eq.~\eqref{eq:cost-aware-objective} is interpretable as nats per unit distance for the
average designer, and the ninefold spread carries the heterogeneity: the two designers
are otherwise identical in design space, observation model, architecture and budget.
Nothing tells a designer which one it is except the rate entering the shared objective, so
any difference in how they behave has to be learned. We compare $\lambda=0$, where cost
is recorded but never penalised, against $\lambda=0.1$.

\begin{figure}[t]
\centering
\includegraphics[width=\linewidth]{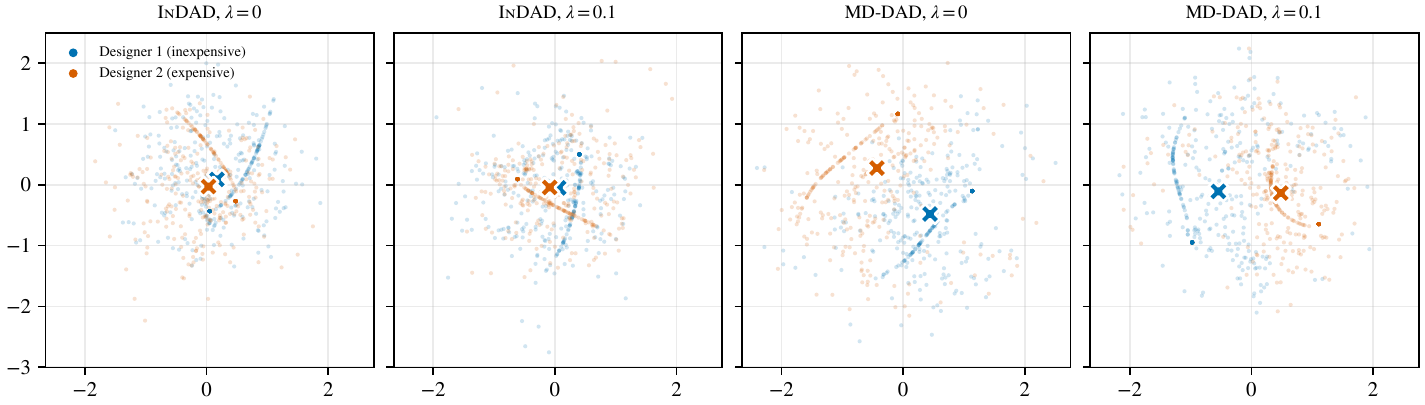}
\caption{\textbf{Measurement locations under heterogeneous costs.} Every measurement of
$64$ rollouts, coloured by designer, with crosses at each designer's mean location. Blue is
the inexpensive designer and orange the expensive one. The crosses pool all rollouts,
whereas the separation in Table~\ref{tab:cost-designs} is measured within each rollout
and then averaged; the two agree for \mddad\ ($1.16$ against $1.33$, $\lambda=0$) because its designers
occupy the same regions in every rollout, and disagree for \indad\ ($0.14$ against
$0.52$, $\lambda=0$ and $\lambda=0.1$) because its designers separate in a direction that varies from rollout to rollout
and averages away.}
\label{fig:cost-designs}
\end{figure}

\paragraph{Results.}
Figure~\ref{fig:cost-designs} shows the
measurement locations, and Table~\ref{tab:cost-designs} repeats Table~\ref{tab:cost}
with the expenditure split between the designers. Two of its columns describe the geometry.
\emph{Separation} is the distance between the two designers' mean measurement locations
within a rollout, which is large when the designers have divided the design space and small
when they are working in the same place, making it a geometric proxy for coordination.
\emph{Spread} is the mean distance of a designer's measurements from its own centre, large
when a designer ranges widely and small when it concentrates.

\begin{table}[t]
\caption{\textbf{Design geometry under heterogeneous costs} (mean $\pm$ s.e.\ over three
seeds). Recreation of Table~\ref{tab:cost} with the expenditure split between the two designers added.}
\label{tab:cost-designs}
\centering\small
\setlength{\tabcolsep}{2.5pt}
\resizebox{\linewidth}{!}{%
\begin{tabular}{lcccccccc}
\toprule
& & & & & \multicolumn{2}{c}{Spread} & \multicolumn{2}{c}{Cost}\\
\cmidrule(lr){6-7}\cmidrule(lr){8-9}
Method & PEIG (lower) & Total cost & Coord.\ gap & Separation & Cheap & Exp. & Cheap & Exp.\\
\midrule
\indad\ ($\lambda=0$)   & $8.34 \pm 0.07$ & $6.83 \pm 0.10$ & $3.11 \pm 0.07$ & $0.52 \pm 0.03$ & $0.55 \pm 0.03$ & $0.53 \pm 0.03$ & $0.69 \pm 0.01$ & $6.15 \pm 0.09$ \\
\indad\ ($\lambda=0.1$) & $8.35 \pm 0.15$ & $6.20 \pm 0.21$ & $3.08 \pm 0.11$ & $0.52 \pm 0.03$ & $0.54 \pm 0.01$ & $0.51 \pm 0.02$ & $0.65 \pm 0.01$ & $5.55 \pm 0.21$ \\
\midrule
\mddad\ ($\lambda=0$)   & $9.56 \pm 0.07$ & $8.34 \pm 0.15$ & $1.25 \pm 0.03$ & $1.33 \pm 0.04$ & $0.64 \pm 0.02$ & $0.70 \pm 0.02$ & $0.79 \pm 0.02$ & $7.55 \pm 0.13$ \\
\mddad\ ($\lambda=0.1$) & $9.55 \pm 0.02$ & $6.49 \pm 0.02$ & $1.32 \pm 0.03$ & $1.23 \pm 0.04$ & $0.69 \pm 0.03$ & $0.61 \pm 0.02$ & $0.80 \pm 0.02$ & $5.69 \pm 0.00$ \\
\bottomrule
\end{tabular}}
\end{table}

\paragraph{Coordination is visible as separation.}
The jointly trained designers keep their centroids $1.33$ apart against $0.52$ for the
independently trained ones, at every cost weight. The figure for independently optimised policies is smaller than
the designers' own spread, which means each designer's measurements lie inside the region its
partner is also measuring: the two are selecting designs with the same part of the design space. \mddad\ instead
partitions the space between its designers (Figure~\ref{fig:cost-designs}).

\paragraph{Costs induce designer specialisation.}
Under the penalty for the jointly trained policy, the expensive designer contracts its spread while the inexpensive designer expands and takes over a
wider exploration (Figure~\ref{fig:cost-designs} and Table~\ref{tab:cost-designs}). The team therefore leaves roaming to the designer that can afford it and
gives the expensive designer a small, tightly clustered set of measurements chosen to be
informative from where it already is. Separation falls minorly, so the designers
remain in distinct regions throughout, and the PEIG is unchanged: the reallocation is free in information terms. \indad, facing the
identical cost structure, contracts both designers slightly, reallocates nothing between
them, and leaves its separation exactly where it was. Cost is thus a second source of
coordination pressure, distinct from redundancy, and the joint objective adapts to it
while the independent one does not.

\section{Decomposing the cost of adding designers}
\label{app:dad-budget}

\begin{figure}[t]
\centering
\includegraphics[width=0.62\linewidth]{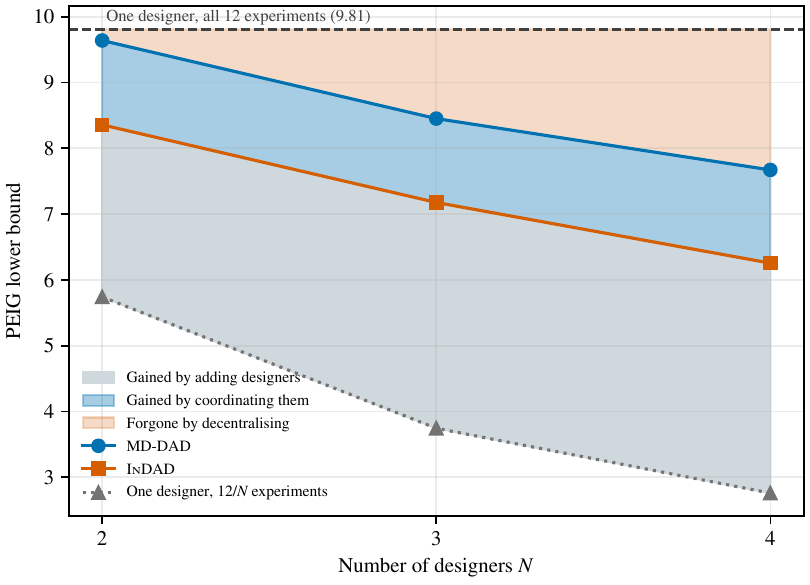}
\caption{\textbf{Adaptivity vs coordination.} 
Four regimes are shown: one
designer restricted to a single member's share of the budget (dotted), $N$ independently
trained designers (\indad), the same $N$ designers trained jointly (\mddad), and one DAD designer
spending the whole budget alone (dashed). The three bands between them are the
quantities this experiment separates. Additional designers increase PEIG
over a designer with the same planning horizon (grey) even when trained independently (\indad): model outcome noise makes otherwise-identical designers diverge. Coordinating with \mddad\ further increases PEIG. The band forgone by decentralising (orange) scales with the
team. The decline in team performance is therefore the loss
of adaptivity under decentralised execution, not the joint objective failing to keep
pace given the blue band is relatively constant compared to the grey band. }
\label{fig:decomposition}
\end{figure}

Section~\ref{subsec:scaling} shows the PEIG of every method falling as the team
grows at a fixed total budget. Two mechanisms could produce that: coordination could be
getting harder, so that the joint objective repairs less of the redundancy, or
adaptivity could be getting weaker, since each designer conditions on a shrinking share of
the data while choosing its designs. Figure~\ref{fig:decomposition} and Table~\ref{tab:dad-budget} separate them by adding
single-designer DAD at each horizon a team member actually experiences.

\begin{table}[t]
\caption{\textbf{Where the information goes as designers are added.} PEIG lower bound
(mean $\pm$ s.e.\ over three seeds) at a fixed total budget of $12$ measurements. The
DAD column is one designer spending only $T$ experiments, so it holds the planning horizon
of a single team member while forgoing the rest of the budget. Per-designer columns score
$I(\theta;H_i)$ (lower) on one designer's own history, exactly as a single-designer bound would, and
average that over the $N$ designers of the team.}
\label{tab:dad-budget}
\centering\small
\setlength{\tabcolsep}{4pt}
\begin{tabular}{llcccccc}
\toprule
& & \multicolumn{1}{c}{One designer} & \multicolumn{2}{c}{Per designer} & \multicolumn{2}{c}{PEIG (lower)} & \\
\cmidrule(lr){3-3}\cmidrule(lr){4-5}\cmidrule(lr){6-7}
$N$ & $T$ & DAD & \indad & \mddad & \indad & \mddad & \mddad$-$\indad\\
\midrule
1 & 12 & $9.81 \pm 0.06$ & n/a & n/a & n/a & n/a & n/a\\
2 & 6 & $5.74 \pm 0.01$ & $5.78$ & $5.42$ & $8.36 \pm 0.08$ & $9.64 \pm 0.09$ & $1.28$\\
3 & 4 & $3.74 \pm 0.04$ & $3.76$ & $3.51$ & $7.18 \pm 0.07$ & $8.45 \pm 0.02$ & $1.28$\\
4 & 3 & $2.76 \pm 0.02$ & $2.75$ & $2.54$ & $6.25 \pm 0.08$ & $7.67 \pm 0.05$ & $1.42$\\
\bottomrule
\end{tabular}
\end{table}

Three readings follow. First, an independently trained designer is a single-designer DAD
policy: its individual bound matches DAD at the same horizon at
every team size.
This is what the independent objective is defined to produce. Jointly trained designers are individually \emph{worse} than this, while their teams are jointly better, which is the coordination gap discussed throughout.

Second, the amount joint training recovers is close to constant in the team size. Coordination failure grows with $N$, since the
coordination gap of \indad\ rises. The decline in team performance is therefore not the joint
objective coping less well as designers are added.

Third, the shortfall against a single designer holding the whole budget grows steadily. Since the coordination component
is flat, this residual is what decentralised execution costs, and it is a loss of
adaptivity rather than of coordination.

\section{Limitations}
\label{app:limitations}
The \framework\ formulation, the PEIG objective and Proposition~\ref{prop:coordination-gap} place no restriction on the likelihood or design space, applying equally to explicit and implicit likelihoods and to continuous and discrete designs. Our empirical evaluation, however, covers only the explicit-likelihood, continuous-design case, since MD-PCE, MD-NMC and \mddad\ require an evaluable likelihood and a reparameterisable simulator. Implicit models and discrete designs could be handled by substituting likelihood-free bounds \citep{ivanova2021implicit} or RL-based policy training \citep{blau2022optimizing}, which we leave to future work.


\clearpage
\section*{NeurIPS Paper Checklist}

\begin{enumerate}

\item {\bf Claims}
    \item[] Question: Do the main claims made in the abstract and introduction accurately reflect the paper's contributions and scope?
    \item[] Answer: \answerYes{}
    \item[] Justification: The abstract and Section~\ref{sec:intro} state the contributions (the \framework\ formulation, the coordination gap, CTDE for BED, and the DAD-based algorithms); these are matched by the results in Sections~\ref{sec:method}--\ref{sec:learning} and the experiments in Section~\ref{sec:experiments}.
   
\item {\bf Limitations}
    \item[] Question: Does the paper discuss the limitations of the work performed by the authors?
    \item[] Answer: \answerYes{}
    \item[] Justification: See Appendix~\ref{app:limitations}.

\item {\bf Theory assumptions and proofs}
    \item[] Question: For each theoretical result, does the paper provide the full set of assumptions and a complete (and correct) proof?
    \item[] Answer: \answerYes{}
    \item[] Justification: Proposition~\ref{prop:coordination-gap} is stated with its assumptions in Section~\ref{subsec:coordination}, and the full proof is given in Appendix~\ref{app:proofs}.
    \item {\bf Experimental result reproducibility}
    \item[] Question: Does the paper fully disclose all the information needed to reproduce the main experimental results of the paper to the extent that it affects the main claims and/or conclusions of the paper (regardless of whether the code and data are provided or not)?
    \item[] Answer: \answerYes{}
    \item[] Justification: Section~\ref{sec:experiments} describes the experiments, and Appendices~\ref{app:estimators} and~\ref{app:details} specify the generative model, estimators, architecture, hyperparameters, number of seeds and evaluation protocol.
\item {\bf Open access to data and code}
    \item[] Question: Does the paper provide open access to the data and code, with sufficient instructions to faithfully reproduce the main experimental results, as described in supplemental material?
    \item[] Answer: \answerNo{}
    \item[] Justification: The results are in tasks which are simulator-based and require no external data, and Appendix~\ref{app:details} provides sufficient detail to reimplement the method. Code is not released with the submission.
    
\item {\bf Experimental setting/details}
    \item[] Question: Does the paper specify all the training and test details (e.g., data splits, hyperparameters, how they were chosen, type of optimizer) necessary to understand the results?
    \item[] Answer: \answerYes{}
    \item[] Justification: Appendix~\ref{app:details} reports the architecture, optimiser, learning rate, gradient steps, inner and outer sample counts, and the experimental budget, all held fixed across methods and team sizes.

\item {\bf Experiment statistical significance}
    \item[] Question: Does the paper report error bars suitably and correctly defined or other appropriate information about the statistical significance of the experiments?
    \item[] Answer: \answerYes{}
    \item[] Justification: All results are averaged over seeds, with error bars denoting $\pm1$ standard error across seeds, as stated in Section~\ref{sec:experiments}. Estimator uncertainty is reported separately as the sPCE/sNMC bracket.

\item {\bf Experiments compute resources}
    \item[] Question: For each experiment, does the paper provide sufficient information on the computer resources (type of compute workers, memory, time of execution) needed to reproduce the experiments?
    \item[] Answer: \answerYes{}
    \item[] Justification: All policy training runs on a single commodity GPU and takes a few hours per run. Per-experiment training settings are reported in Appendix~\ref{app:details}.
    
\item {\bf Code of ethics}
    \item[] Question: Does the research conducted in the paper conform, in every respect, with the NeurIPS Code of Ethics \url{https://neurips.cc/public/EthicsGuidelines}?
    \item[] Answer: \answerYes{}
    \item[] Justification: The research conforms with the NeurIPS Code of Ethics; it does not involve human subjects or personal data, and uses only simulator-based benchmarks.

\item {\bf Broader impacts}
    \item[] Question: Does the paper discuss both potential positive societal impacts and negative societal impacts of the work performed?
    \item[] Answer: \answerYes{}
    \item[] Justification: The work concerns how experimental effort is allocated among designers, so its impacts follow those of the experiments it coordinates: acquiring the same information from fewer experiments reduces cost, time and burden on participants, while the same efficiency is available to campaigns whose purpose is harmful. The paper releases no data or models that could be repurposed directly.
    
\item {\bf Safeguards}
    \item[] Question: Does the paper describe safeguards that have been put in place for responsible release of data or models that have a high risk for misuse (e.g., pre-trained language models, image generators, or scraped datasets)?
    \item[] Answer: \answerNA{}
    \item[] Justification: The paper does not release any pretrained models, generative models or scraped datasets; it introduces a design objective evaluated on simulators.

\item {\bf New assets}
    \item[] Question: Are new assets introduced in the paper well documented and is the documentation provided alongside the assets?
    \item[] Answer: \answerNA{}
    \item[] Justification: The paper does not release new datasets or pretrained models. The proposed method is documented in Sections~\ref{sec:method}--\ref{sec:learning} and Appendix~\ref{app:details}.
   
\item {\bf Crowdsourcing and research with human subjects}
    \item[] Question: For crowdsourcing experiments and research with human subjects, does the paper include the full text of instructions given to participants and screenshots, if applicable, as well as details about compensation (if any)? 
    \item[] Answer: \answerNA{}
    \item[] Justification: The paper does not involve crowdsourcing or research with human subjects.

\item {\bf Institutional review board (IRB) approvals or equivalent for research with human subjects}
    \item[] Question: Does the paper describe potential risks incurred by study participants, whether such risks were disclosed to the subjects, and whether Institutional Review Board (IRB) approvals (or an equivalent approval/review based on the requirements of your country or institution) were obtained?
    \item[] Answer: \answerNA{}
    \item[] Justification: The paper does not involve research with human subjects, so no IRB approval was required.
\item {\bf Declaration of LLM usage}
    \item[] Question: Does the paper describe the usage of LLMs if it is an important, original, or non-standard component of the core methods in this research? Note that if the LLM is used only for writing, editing, or formatting purposes and does \emph{not} impact the core methodology, scientific rigor, or originality of the research, declaration is not required.
    \item[] Answer: \answerNA{}
    \item[] Justification: LLMs are not used as part of the core methodology of this research.

\end{enumerate}

\end{document}